\documentclass[10pt,twocolumn,letterpaper]{article}

\usepackage{cvpr}              
\usepackage{multirow} 

\usepackage{adjustbox}
\definecolor{cvprblue}{rgb}{0.21,0.49,0.74}
\usepackage[pagebackref,breaklinks,colorlinks,allcolors=cvprblue]{hyperref}

\def\paperID{} 
\def\confName{CVPR}
\def\confYear{2026}

\title{OMG-Avatar: One-shot Multi-LOD Gaussian Head Avatar}

\author{Jianqiang Ren,~~Lin Liu,~~Steven Hoi \\ 
Tongyi Lab,~~Alibaba Group\\
}

\begin{document}
\maketitle
\begin{abstract}
We propose \textbf{OMG-Avatar}, a novel \textbf{O}ne-shot method that leverages a \textbf{M}ulti-LOD (Level-of-Detail) \textbf{G}aussian representation for animatable 3D head reconstruction from a single image in 0.2s. Our method enables LOD head avatar modeling using a unified model that accommodates diverse hardware capabilities and inference speed requirements. To capture both global and local facial characteristics, we employ a transformer-based architecture for global feature extraction and projection-based sampling for local feature acquisition. These features are effectively fused under the guidance of a depth buffer, ensuring occlusion plausibility. We further introduce a coarse-to-fine learning paradigm to support Level-of-Detail functionality and enhance the perception of hierarchical details. To address the limitations of 3DMMs in modeling non-head regions such as the shoulders, we introduce a multi-region decomposition scheme in which the head and shoulders are predicted separately and then integrated through cross-region combination. Extensive experiments demonstrate that OMG-Avatar outperforms state-of-the-art methods in reconstruction quality, reenactment performance, and computational efficiency.
\end{abstract}    
\section{Introduction}
\label{sec:intro}
Reconstructing an animatable 3D head avatar from a single image is a crucial and rapidly evolving research area in computer vision and graphics. This technology has great potential for applications across various domains, including the game and video production industries, virtual meetings, and the emerging Metaverse. To facilitate the widespread adoption of this technology, several key features are essential: high-efficiency reconstruction and inference, rich facial details, and precise controllability over expressions and head poses. In recent years, numerous methods have been developed to tackle this task, which can be divided into 2D-based and 3D-based approaches.

Early 2D-based methods~\cite{siarohin2019first,wang2021one,guo2024liveportrait} predict deformation flows to warp the latent features of a source portrait and employ GANs (Generative Adversarial Networks)~\cite{goodfellow2014generative} to synthesize the reenacted output. With the rise of latent diffusion models~\cite{rombach2022high}, recent approaches~\cite{tian2024emo, jiang2024loopy, zhao2025x} have adopted cross-attention mechanisms conditioned on driving signals, achieving superior image quality and better appearance preservation. However, both GAN-based and diffusion-based methods require substantial computational resources, limiting their applicability in real-time scenarios. Furthermore, due to the lack of 3D constraints, these approaches often struggle to maintain multi-view consistency under large pose or viewpoint variations.

In the 3D avatar domain, NeRF and Gaussian Splatting have emerged as prominent approaches due to their high-quality representation and rendering capabilities. Compared to NeRF-based methods~\cite{bai2023high,gafni2021dynamic,ki2024learning, li2023one, ma2023otavatar, park2021nerfies, yu2023nofa, zheng2023pointavatar}, 3D Gaussian Splatting has become the prevailing choice owing to its significantly fast rendering speed. Unlike approaches that require extensive per-individual optimization or multi-view inputs~\cite{xu2024gaussian, wu2024gaussian, tang2025gaf, zheng2025headgap, xiang2024flashavatar}, recent methods such as~\cite{chu2024generalizable, he2025lam, kirschstein2025avat3r,zhang2025guava} generate 3D avatars from a single image, significantly enhancing generalization capabilities. 
Despite these advancements, key challenges persist in efficiency, computational cost, and compatibility. For instance, GAGAvatar~\cite{chu2024generalizable} primarily utilizes Gaussian points sampled from the dual-lifting planes of backgrounds, leading to redundancy and inefficiency. 
LAM~\cite{he2025lam} employs subdivided vertices as queries in its cross-attention mechanism, and its computational complexity grows exponentially with the number of subdivision levels. Avat3r~\cite{kirschstein2025avat3r} relies on an additional 3D GAN for 3D lifting to enable single-image reconstruction, which introduces error accumulation. Moreover, they all fail to dynamically adjust computational complexity to accommodate diverse hardware capabilities and inference speed requirements.

To address these challenges, we propose a novel one-shot method for generating multi-LOD 3D Gaussian head avatars using a unified model. Rather than directly performing 2D-to-3D feature mapping on high-resolution meshes, which is computationally expensive, our approach extracts both global and local features from low-resolution meshes and progressively refines them to high-resolution representations through subdivision operations during the training phase, which also enables dynamic multi-LOD rendering at runtime. By leveraging an effective occlusion-aware feature fusion mechanism, our model delivers superior reconstruction quality while significantly reducing computational cost and the number of Gaussians compared to existing methods. Additionally, to improve representation in non-head regions, we independently model the head and shoulder based on shared features, improving the completeness of the generated avatars.

The main contributions are summarized as follows:
\begin{itemize}
\item We present OMG-Avatar, a novel framework that reconstructs animatable Gaussian head avatars from a single image in just 0.2 seconds. It enables dynamic multi-level-of-detail (multi-LOD) rendering while achieving state-of-the-art reconstruction quality and real-time inference at 85 FPS using fewer Gaussians than prior methods.

\item We propose a coarse-to-fine learning strategy that progressively refines both transformer-based global features and projection-sampled local features through multi-level subdivisions, followed by an occlusion-aware feature fusion mechanism guided by the depth buffer. These components significantly enhance the fidelity and completeness of the generated avatars.

\item Comprehensive experiments on two large-scale datasets demonstrate that OMG-Avatar surpasses state-of-the-art methods in reconstruction accuracy, reenactment fidelity, and computational efficiency.

\end{itemize}
\section{Related Work}
\label{sec:related_work}

\noindent{}{\bf 2D Talking Head Generation.} Early 2D approaches~\cite{zakharov2019few, burkov2020neural, zhou2021pose, wang2023progressive} employ generative adversarial networks (GANs)~\cite{goodfellow2014generative, isola2017image, karras2020analyzing} and incorporate driving expression features for controllable portrait synthesis. Subsequent methods~\cite{siarohin2019first, ren2021pirenderer, drobyshev2022megaportraits, hong2022depth, zhang2023metaportrait, guo2024liveportrait} adopt deformation-based frameworks, representing expressions and poses as warping fields to deform the source image. These methods usually struggle with large pose and expression variations due to a lack of 3D awareness. To address this, some~\cite{nef1999morphable, paysan20093d, li2017learning, gerig2018morphable} integrate 3D Morphable Models (3DMMs) into 2D pipelines, but they still lack support for viewpoint control and free-viewpoint rendering. Recent diffusion-based approaches~\cite{cui2024hallo2, tian2024emo, xu2024hallo, zhao2025x} further improve visual quality and spatio-temporal coherence. However, their high computational cost hinders real-time performance, such as in video conferencing or live chat applications.

\noindent{}{\bf 3D Head Avatar Generation.} Traditional 3D head avatars typically rely on 3D Morphable Models (3DMMs) for mesh reconstruction~\cite{xu2020deep, khakhulin2022realistic}, which often fail to capture fine geometric details. In contrast, NeRF-based approaches~\cite{kirschstein2023nersemble, athar2022rignerf, bai2023learning, gafni2021dynamic, gao2022reconstructing, guo2021ad, ki2024learning, park2021nerfies, park2021hypernerf, tretschk2021non, zhang2024learning, zhao2023havatar, zheng2023pointavatar, zielonka2023instant} have significantly improved reconstruction accuracy and detail representation, and several efficient one-shot NeRF methods have also been proposed~\cite{yu2023nofa, li2023one, yang2024learning, chu2024gpavatar, ma2024cvthead}. However, these NeRF-based methods still face challenges in achieving real-time rendering performance.
This limitation is addressed by 3D Gaussian Splatting~\cite{kerbl20233d}, which offers faster rendering while maintaining high visual quality. Unlike earlier Gaussian-based methods~\cite{xu2024gaussian, wu2024gaussian, tang2025gaf, zheng2025headgap} that rely on extensive individual-specific optimization or multi-view inputs, recent works~\cite{chu2024generalizable, he2025lam, kirschstein2025avat3r,zhang2025guava} propose one-shot reconstruction frameworks for head avatars. However, these methods still face challenges such as redundant Gaussian utilization, excessive computational demands, and incomplete head representation, which we aim to address in this paper.

\noindent{}{\bf Hierarchical Gaussian Representation.} The hierarchical representation has been widely adopted for efficiently modeling multi-scale and structured data. HiSplat~\cite{tang2024hisplat} introduces a hierarchical Gaussian splatting framework for sparse-view reconstruction, employing coarse-grained Gaussians to capture large-scale geometry and fine-grained Gaussians to represent texture details. LODAvatar~\cite{dongye2024lodavatar} integrates level-of-detail (LOD) control into Gaussian avatars through hierarchical embedding, achieving a favorable trade-off between visual quality and computational cost. However, it does not support facial animation. GaussianHeads~\cite{teotia2024gaussianheads} leverages a hierarchical representation to model complex facial expressions and head dynamics, but requires over 20 hours of training and relies on a multi-view camera rig. In contrast, our work employs a hierarchical representation to enable both efficient modeling and dynamic LOD rendering at runtime, supporting real-time, animatable avatars without the need for specialized capture setups.

\section{Method}
\label{sec:method}
\begin{figure*}[ht]
\centering
\includegraphics[width=0.99\linewidth]{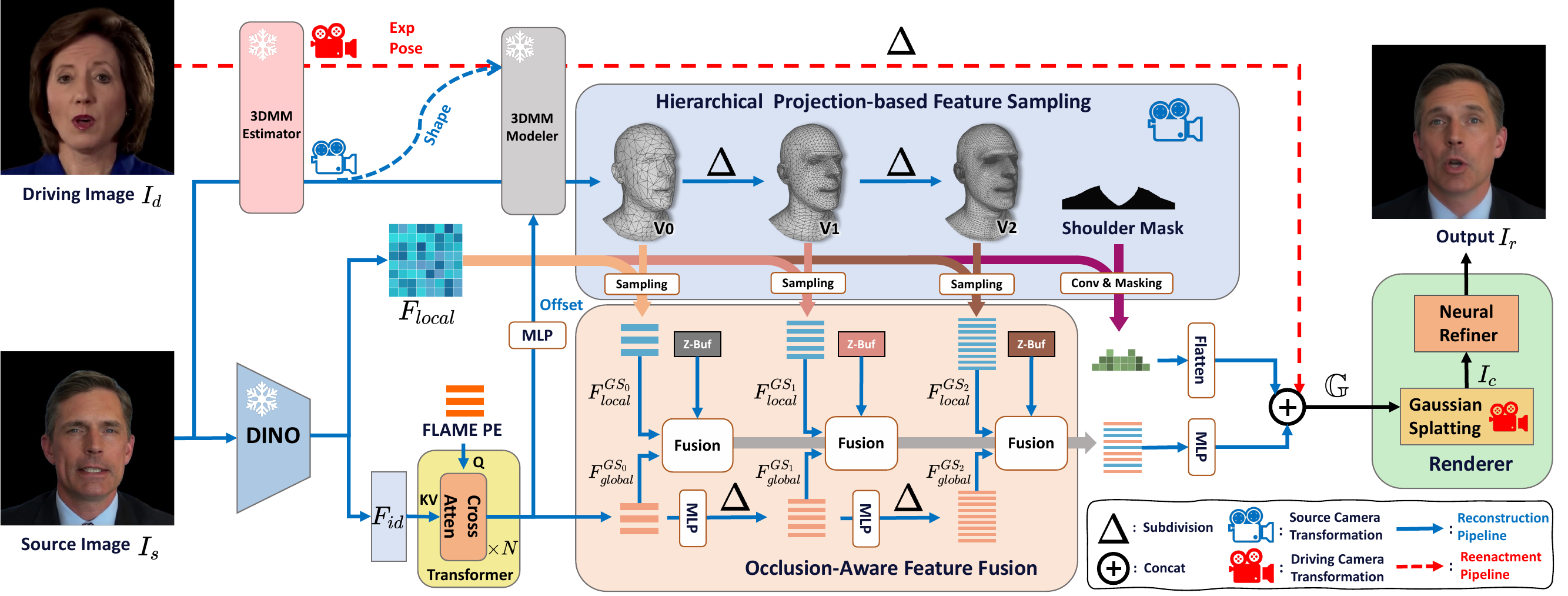}
\caption{The overall pipeline of OMG-Avatar framework. Our method extracts global features via cross-attention and local details via projection-based sampling, which are fused under the guidance of depth buffers. A coarse-to-fine strategy is proposed to facilitate hierarchical detail perception. The head and shoulder are predicted separately using shared features and then combined for rendering.}
\vspace{-5pt}
\label{fig: architecture}
\end{figure*}

~\cref{fig: architecture} illustrates the overall pipeline of our method. Given a source image, we first extract both local and identity features using DINOv2~\cite{oquab2023dinov2} and estimate the 3D head mesh via a 3DMM modeler. The Hierarchical Projection-based Feature Sampling (HPFS) module then projects the head mesh onto the image plane to sample local features at the corresponding coordinates. Concurrently, the global feature is obtained through cross-attention, where FLAME positional embeddings serve as queries. 
Subsequently, the Occlusion-Aware Feature Fusion(OAFF) module fuses the global and local features under the guidance of depth buffers, ensuring spatial coherence and occlusion plausibility. Features from non-head regions are further integrated with head-related features to jointly predict Gaussian attributes for splatting rendering. Finally, a neural renderer generates a refined output image based on the coarse splatted feature maps.
During training, we progressively subdivide the meshes along with their corresponding features, enabling the network to capture hierarchical details in a coarse-to-fine manner, which enhances both training stability and reconstruction accuracy. The details of each module are explained in the subsequent sections.

\subsection{Hierarchical Global-Local Feature Extraction}
\begin{figure}[ht]
\centering
\includegraphics[width=0.99\linewidth]{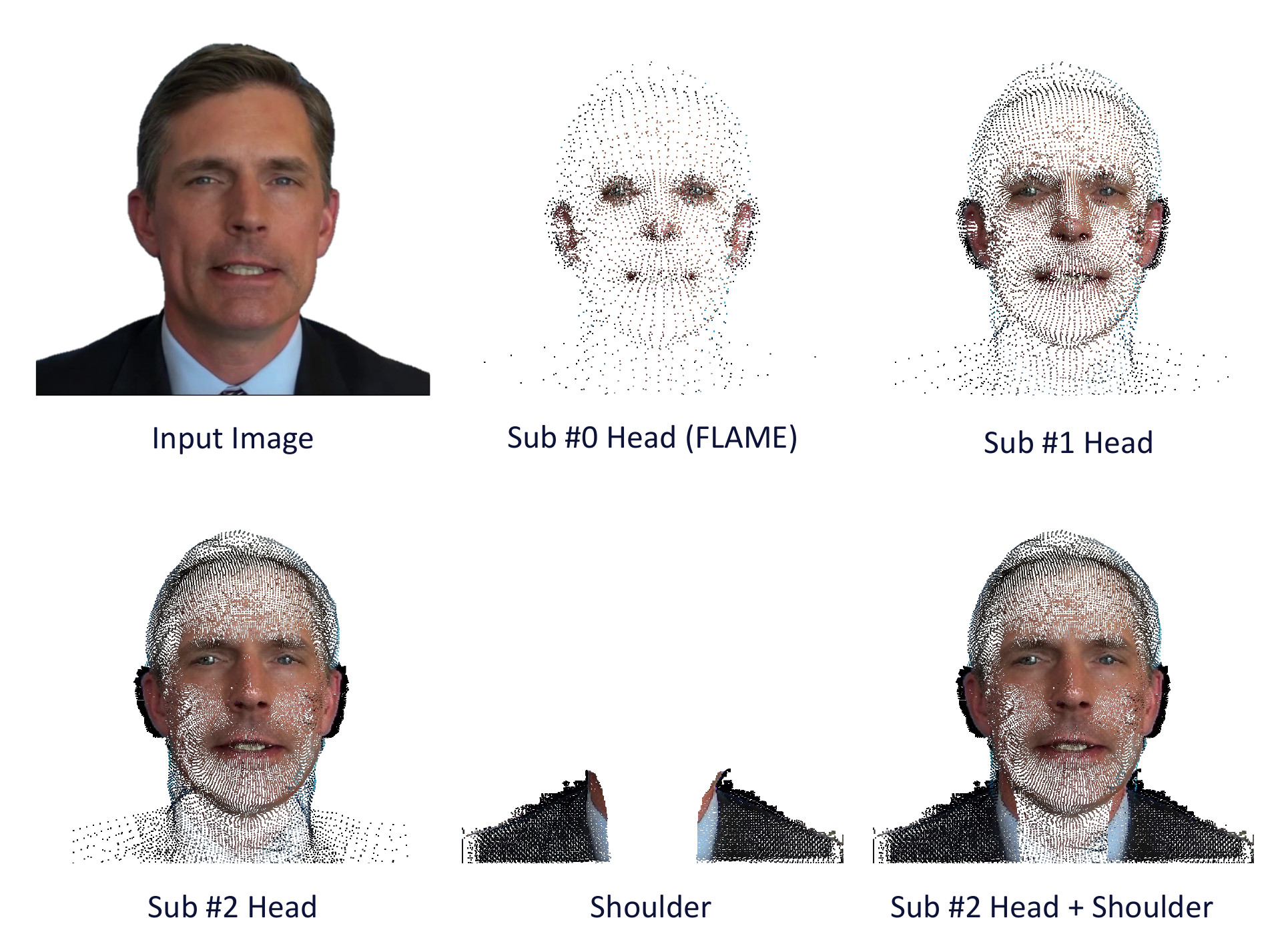}
\vspace{-5pt}
\caption{Since the original FLAME model lacks shoulder regions and sufficient geometric representation for high-resolution LOD requirements, we progressively subdivide the head mesh during training and enhance it with predicted shoulder geometry through cross-region combination. Sub \#$i$ indicates that the head mesh has been subdivided $i$ times.
}
\label{fig: subdivision}
\vspace{-5pt}
\end{figure}

For the task of generating a 3D head model from a single image, the core objective is to establish a 2D-to-3D feature mapping mechanism that transforms image features into 3D spatial features. 
To incorporate statistical priors on head geometry, we employ the FLAME~\cite{li2017learning} model as the 3D head representation, which comprises $N_{0}$ = 5023 vertices. We leverage DINOv2~\cite{oquab2023dinov2} to extract both local features $F_{local}$ and identity features $F_{id}$ from the source image $I_{s}$ following~\cite{chu2024generalizable}. For the $F_{id}$, we assign a learnable positional encoding to each vertex of FLAME as a query, and employ multiple cross-attention blocks to extract global features $F^{GS_{0}}_{global}$. The $F^{GS_{0}}_{global}$ is then used to predict vertex offsets via an MLP $\Phi_{\text{offset}}$ to improve the precision  of the estimated head mesh $T_p$ via a 3DMM modeler:
\begin{equation} \label{eq: flame}
\begin{split}
T_p(\vec{\beta},\vec{\theta},\vec{\psi } )= & \overline{T} +B_S( \vec{\beta}  ;S)+B_P(\vec{\theta};P) \\ & +B_E(\vec{\psi };E)+  \Phi _{\text{offset}}(F^{GS_0}_{global}) ,
\end{split}
\end{equation}
where $\overline{T}$ is the template mesh, and $B_S$, $B_P$ and $B_E$ represent shape, pose, and expression blendshapes respectively. The initial head vertices $V_0$ are obtained using a standard skinning function $W$:
\begin{equation} \label{eq: flame}
V_0=W(T_p(\vec{\beta},\vec{\theta},\vec{\psi }),\boldsymbol{J}(\vec{\beta}),\vec{\theta},\boldsymbol{\mathcal{W}}),
\end{equation}
where $W$ rotates the morphed mesh $T_p$ around joints $\boldsymbol{J}$ and smooths it using blendweights $\boldsymbol{\mathcal{W}}$. 
We project $V_{0}$ into the image space to obtain the corresponding pixel coordinates for each vertex and perform bilinear sampling on $F_{local}$ to extract per-vertex features, denoted as $F^{GS_{0}}_{local}$. As noted in LAM~\cite{he2025lam}, the original number of $V_{0}$ is insufficient for detailed modeling, so we introduce a coarse-to-fine strategy that progressively subdivides the mesh $V_{k}$ and its associated global features $F^{GS_{k}}_{global}$ during training:
\begin{equation} \label{eq: subdivider}
F^{GS_{k+1}}_{global}, V_{k+1} =  \Delta(\Phi_{k}( F^{GS_{k}}_{global}), V_{k}),\quad 0\le k \le K,
\end{equation}
where $\Delta$ denotes the mesh subdivision operation, $\Phi_{k}$ is an MLP network, and $k$ indicates the subdivision level. With the refined vertices $V_{k}$, the corresponding local feature can be obtained as:
\begin{equation} \label{eq: sampling}
F^{GS_{k}}_{local}= Sampling(\mathrm{P}(V_{k}), F_{local}) ,\quad 0\le k \le K,
\end{equation}
where $\mathrm{P}$ is the camera projective transformation (perspective camera with a focal length of 12). As $k$ increases, the resolution of head meshes and features are continuously refined. ~\cref{fig: subdivision} shows the subdivided vertices. To balance quality and efficiency, we set the maximum subdivision level to $K$ = 2, resulting in 79, 936 vertices. Notably, unlike LAM which performs costly cross-attention across all 80K vertices, we computes cross-attention at the initial level with $N_{0}$ = 5K vertices. High-resolution geometry is incrementally refined through efficient subdivision and sampling, significantly reducing computational and memory costs while maintaining reconstruction quality.

\subsection{Occlusion-Aware Feature Fusion}
Once hierarchical global and local features are extracted, we propose an occlusion-aware fusion strategy guided by the depth buffer to ensure robust feature integration. During the rasterization of the 3D mesh, occlusion culling is applied to invisible vertices. Consequently, local features sampled via projection are accurate for visible vertices but may be ambiguous for occluded ones, as their corresponding 2D image features are absent. On the other hand, the global features contain high-level semantic information from the input image, allowing them to infer plausible representations for occluded regions but lacking high-frequency details. Building upon this observation, we leverage the depth buffer to identify and retain only high-confidence local features from visible vertices, selectively fusing them with global features. Specifically, when each vertex is projected into the image space, its depth value is compared with the depth buffer at the corresponding pixel location. Based on this comparison, a binary visibility mask $ M^{GS_{k}} \in \{0, 1\}^{N_{k}} $ is constructed. Formally, for each vertex $ v_i \in V_{k} $, let $ z_i $ be its depth in camera space, and $ \hat{z}_i $ be the depth value recorded in the depth buffer. The mask $ M^{GS_{k}} \in \{0, 1\}^{N_{k}} $  is defined as:
\begin{equation}
M^{GS_{k}}_i =
\begin{cases}
1, & \text{if } z_i = \hat{z}_i \\
0, & \text{if } z_i > \hat{z}_i
\end{cases}, \quad \forall~i = 1, 2, \dots, N_{k},
\end{equation}
Here, $ M^{GS_{k}}_i$ = 1 indicates that the vertex $ v_i $ is visible, while $ M^{GS_{k}}_i$ = 0 identifies occluded vertices. The final fused head feature $F^{GS_{k}}_{h}$ is computed as: 
\begin{equation} \label{eq: ZbuffFusion}
F^{GS_{k}}_{h}= F^{GS_{k}}_{global} + F^{GS_{k}}_{local} \odot  M^{GS_{k}},\quad 0\le k \le K.
\end{equation}
By effectively combining the strengths of global and local features, we achieve a more robust representation of the head vertices. A set of MLPs $\Phi$ are then employed to regress the Gaussian attributes for each head vertex, including color $c_{h}$, opacity $o_{h}$, scale $s_{h}$, and rotation $r_{h}$:
\begin{equation} \label{eq: HeadGS}
c_{h},o_{h},s_{h},r_{h} =  \Phi_{c,o,s,r} (F^{GS_{K}}_{h}).
\end{equation}
The positions $p_h$ are directly derived from the subdivided mesh $V_{K}$, resulting in a full set of Gaussian parameters $\mathbb{H}$ that describe the head model:
\begin{equation} \label{eq: HeadGS}
\mathbb{H}=\{c_{h},o_{h},s_{h},r_{h},p_{h}=V_{K}\}.
\end{equation}

\subsection{Multi-region Modeling and Integration}
The FLAME model lacks sufficient vertex coverage in the shoulder region, resulting in coarse and blurry reconstructions in prior methods. To address this limitation, inspired by~\cite{wu2024gaussian}, we perform image segmentation on the source image to obtain a shoulder mask $M_s$. The extracted local features $F_{local}$ are passed through a convolutional neural network to generate a feature plane, where separate channels encode the Gaussian rendering attributes. The shoulder-relevant region is then isolated using the mask, obtaining the corresponding parameters as follows:
\begin{equation} \label{eq: shoulder-attributes}
c_s, o_s, s_s, r_s, O_s = \mathrm{Flatten}(\mathrm{Conv}(F^{GS}_{local}) \odot M_{s}),
\end{equation}
where $O_s$ represents the offset for shoulder points. For position estimation, we generate the 3D shoulder points $\hat{p}_s$ on an image-aligned plane in world space based on the given camera transformation and feature plane resolution. These points are paired with their corresponding direction vectors $n_s$. The final shoulder points are then calculated as:
\begin{equation} \label{eq: shoulder-points}
\mathbb{S} = \{c_{s}, o_{s}, s_{s}, r_{s}, p_{s} = \hat{p}_s + O_s \cdot n_s\}.
\end{equation}
Finally, we concatenate the head and shoulder parameter sets along the attribute dimensions to form a complete Gaussian parameter set $\mathbb{G}$ that covers both head and shoulder regions:
\begin{equation} \label{eq: ZbuffFusion}
 \mathbb{G}=  \mathbb{H} \oplus  \mathbb{S}.
\end{equation}

\subsection{Reenactment and Rendering}
After reconstructing the Gaussian head avatar $ \mathbb{G}$, our framework enables efficient reenactment, allowing the model to mimic facial expressions and head movements observed in a target video. As shown in ~\cref{fig: architecture}, given a driving image $ I_d $, the expression and pose parameters are extracted using a 3DMM estimator. These driving parameters are then combined with the identity-related parameters derived from $ I_s$ to generate novel FLAME vertices. The FLAME vertices are further refined via subdivisions, resulting in the final head positions $ p_h $ used for Gaussian rendering. Notably, the reenactment process only needs to update the positional component $ p_h $ in the Gaussian parameter set $ \mathbb{G} $ (the red dashed line in ~\cref{fig: architecture}), enabling real-time rendering of avatar animations. To enhance the expressiveness of the Gaussian representation, we adopt a rendering pipeline inspired by~\cite{chu2024generalizable}. Specifically, instead of rendering RGB values, we predict multi-channel feature maps, where the first three channels encode a coarse RGB image $I_c$. The feature maps are subsequently refined by a UNet-based neural refiner to produce the final high-quality output $I_r$.

\subsection{Learning Strategy}
The training process is conducted on a large-scale human video dataset in a self-supervised manner. For each video, we randomly sample two frames and assign one as the source image $ I_s $ and the other as the driving image $ I_d $. The objective is to train the network to generate an output image that closely resembles the driving image in both appearance and motion.

To achieve this, we employ a multi-component loss function, combining L2 loss, SSIM loss, and perceptual loss, applied to both the coarse and refined output images. Additionally, to constrain the displacement of FLAME model vertices, we impose a regularization term on the vertex offsets. The total loss is defined as:
\begin{equation}
\begin{split}
\mathcal{L} = & \lambda_1 \mathcal{L}_{\text{2}}(I_d, I_c \& I_r) + \lambda_2 \mathcal{L}_{\text{SSIM}}(I_d, I_c \& I_r) \\
& + \lambda_3 \mathcal{L}_{\text{percep}}(I_d, I_c \& I_r) + \lambda_4 \mathcal{L}_{\text{reg}},
\end{split}
\end{equation}
where $\mathcal{L}(A, B \& C)$ denotes $\mathcal{L}(A, B)+ \mathcal{L}(A, C)$, $\mathcal{L}_{\text{2}}$, $\mathcal{L}_{\text{SSIM}}$, and $\mathcal{L}_{\text{percep}}$ are computed on both the $I_c$ and refined $I_r$, and $\mathcal{L}_{\text{reg}} =  \left \| \text{offset} \right \|_2 $ penalizes large vertex displacements.

\section{Experiment}
\subsection{Datasets and Settings}
\noindent {\bf Datasets.}
Our model is trained on the VFHQ~\cite{xie2022vfhq} dataset, which contains video clips from a variety of interview scenarios. To ensure temporal diversity, we uniformly sample frames from each video following previous works~\cite{chu2024generalizable,he2025lam}, leading to a total of 766, 263 frames across 15, 204 video clips. All images are cropped to focus on the head region, resized to $ 512 \times 512 $ pixels for consistency, and further processed with camera pose tracking, FLAME parameter estimation, and background removal as described in prior works~\cite{chu2024generalizable,chu2024gpavatar}. For evaluation, we adopt the official test split of the VFHQ dataset, comprising 2, 500 frames from 50 videos. The first frame of each video is used as the source image, while the remaining frames serve as driving and target images for reenactment. Additionally, we evaluate our method on the HDTF dataset~\cite{zhang2021flow} using the model trained solely on VFHQ to demonstrate its generalization capability.
We process the HDTF dataset using the standard test split introduced in~\citet{ma2023otavatar} and~\citet{li2023generalizable}, which includes 19 video sequences. 

\noindent{}{\bf Implementation Details.}
Our transformer network for extracting global features consists of two decoder layers, each with 8 attention heads. The dimension of the FLAME positional encoding is set to 256. Rather than relying on an external semantic segmentation model, we derive the shoulder region mask by calculating the difference between the portrait mask at the bottom of the image and the depth buffer mask in a straightforward manner. The average number of Gaussian points in the shoulder region is 9K. During training, the weights of the DINOv2 and 3DMM estimator modules are frozen. We train the entire model on a single NVIDIA A100 GPU for 6 epochs using the Adam optimizer with a learning rate of $1 \times 10^{-4}$ and a batch size of 8. The subdivide levels are gradually increased based on the training stage.
We set the loss parameters $\lambda_1=10$, $\lambda_2 = 1$, and $\lambda_3 = \lambda_4 = 0.1$. More details are provided in the supplemental material.

\noindent{}{\bf Evaluation Metrics.}
To comprehensively evaluate the performance of both self- and cross-identity reenactment, we employ a multi-faceted assessment framework that incorporates a variety of quantitative metrics. For self-reenactment scenarios where ground-truth data is available, we assess the quality of generated images using three widely adopted objective measures: Peak Signal-to-Noise Ratio (PSNR), Structural Similarity Index (SSIM), and Learned Perceptual Image Patch Similarity (LPIPS)~\cite{zhang2018perceptual}. These metrics provide reliable comparisons between synthesized outputs and reference ground-truth images. To evaluate identity preservation, we compute the cosine distance between face recognition features extracted from the source and reenacted images, following the methodology proposed in~\citet{deng2019arcface}. For assessing the accuracy of expression and pose transfer, we utilize a 3D Morphable Model (3DMM) estimator~\cite{deng2019accurate} to calculate the Average Expression Distance (AED) and Average Pose Distance (APD). Additionally, we use facial landmark detection~\cite{bulat2017far} to measure the Average Keypoint Distance (AKD), which provides further insight into the precision of motion control during animation. In the case of cross-identity reenactment, where ground-truth data is not available, we adopt an evaluation protocol based on Consistency of Identity Similarity (CSIM), AED, and APD—metrics aligned with those used in recent studies~\cite{chu2024generalizable, he2025lam}. This ensures comparability across different methods and enables meaningful analysis.

\subsection{Baseline methods}
We conduct a comprehensive comparison between our method and state-of-the-art 3D avatar approaches, including ROME~\cite{khakhulin2022realistic}, StyleHeat~\cite{yin2022styleheat}, OTAvatar~\cite{ma2023otavatar}, HideNeRF~\cite{li2023one}, GOHA~\cite{li2023generalizable}, CVTHead~\cite{ma2024cvthead}, GPAvatar~\cite{chu2024gpavatar}, Real3DPortrait~\cite{ye2024real3d}, Portrait4D~\cite{deng2024portrait4d}, Portrait4D-v2~\cite{deng2024portrait4dv2}, GAGAvatar~\cite{chu2024generalizable}, and LAM~\cite{he2025lam}. For each baseline, we utilize its official implementation to produce results. For our model, the level-of-detail (LOD) can be controlled at inference time, Sub \#$i$ indicates that the head mesh has been subdivided $i$ times.


\begin{figure*}[t]
\begin{center}
\adjustbox{trim=0 0.00\height{} 0 0, clip}{
\includegraphics[width=0.99\linewidth]{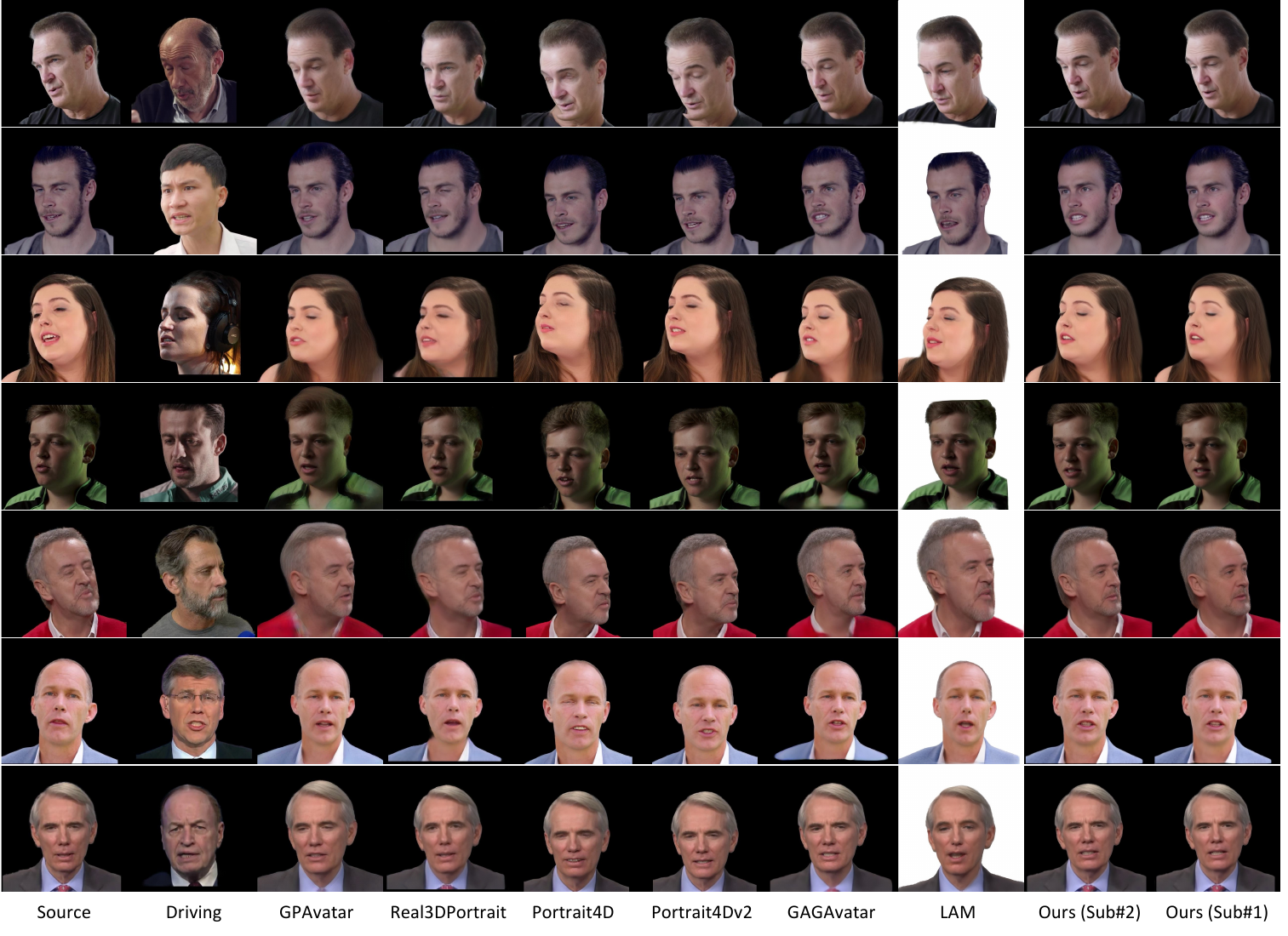}
}
\end{center}
\vspace{-10pt}
\caption{Cross-identity reenactment results on VFHQ and HDTF datasets.}
\vspace{-10pt}
\label{fig: cmp}
\end{figure*}

\subsection{Qualitative Evaluation}
We compare our method against baseline approaches on the VFHQ and HDTF datasets, with qualitative results shown in~\cref{fig: cmp}. As can be seen from the figure, GPAvatar~\cite{chu2024gpavatar} suffers from inaccurate expression tracking and exhibits noticeable blurring in the neck region—a limitation also observed in GAGAvatar~\cite{chu2024generalizable} (see rows 4 and 5). Although Real3DPortrait~\cite{ye2024real3d}, Portrait4D~\cite{deng2024portrait4d}, and Portrait4Dv2~\cite{deng2024portrait4dv2} achieve high visual fidelity, they introduce severe misalignment artifacts under certain poses, manifesting as prominent cracking near the neck and chin (row 1). Additionally, LAM~\cite{he2025lam} exhibits unnatural shoulder tilting (rows 1 and 3) alongside noticeable artifacts around the mouth and teeth (rows 2 and 6).
In contrast, our method achieves superior visual quality compared to existing approaches while using significantly fewer Gaussian points. Notably, even our low-resolution variant (Sub \#1 with $\sim$29K Gaussian points), shown in the last column of~\cref{fig: cmp}, maintains comparable visual fidelity, making it particularly well-suited for deployment in high-speed applications or on hardware with limited computational resources.


\begin{table*}[t]
\centering
\caption{Quantitative results on the VFHQ dataset. The
\colorbox{red!40}{first}, \colorbox{orange!40}{second}, and \colorbox{yellow!40}{third} best-performing methods are highlighted. The Sub \# indicates the subdivision level for inference.}
\label{tab:vfhq}
\resizebox{0.9\textwidth}{!}{
\begin{tabular}{l|ccccccc|ccc}
\toprule
\multirow{2}{*}{\textbf{Method}}
& \multicolumn{7}{c|}{\textbf{Self Reenactment}}
& \multicolumn{3}{c}{\textbf{Cross Reenactment}} \\
& \textbf{PSNR}$\uparrow$ & \textbf{SSIM}$\uparrow$
& \textbf{LPIPS}$\downarrow$ & \textbf{CSIM}$\uparrow$
& \textbf{AED}$\downarrow$ & \textbf{APD}$\downarrow$ & \textbf{AKD}$\downarrow$
& \textbf{CSIM}$\uparrow$ & \textbf{AED}$\downarrow$ & \textbf{APD}$\downarrow$ \\
\midrule
StyleHeat        & 19.95 & 0.726 & 0.211 & 0.537 & 0.199 & 0.385 & 7.659 & 0.407 & 0.279 & 0.551 \\
ROME                 & 19.96 & 0.786 & 0.192 & 0.701 & 0.138 & 0.186 & 4.986 & 0.530 & 0.259 & 0.277 \\
OTAvatar         & 17.65 & 0.563 & 0.294 & 0.465 & 0.234 & 0.545 & 18.19 & 0.364 & 0.324 & 0.678 \\
HideNeRF          & 19.79 & 0.768 & 0.180 & 0.787 & 0.143 & 0.361 & 7.254 & 0.514 & 0.277 & 0.527 \\
GOHA                 & 20.15 & 0.770 & 0.149 & 0.664 & 0.176 & 0.173 & 6.272 & 0.518 & 0.274 &  0.261 \\
CVTHead            & 18.43 & 0.706 & 0.317 & 0.504 & 0.186 & 0.224 & 5.678 & 0.374 & 0.261 & 0.311 \\
GPAvatar           & 21.04 & 0.807 & 0.150 & 0.772 & 0.132 & 0.189 & 4.226 & 0.564 & 0.255 & 0.328 \\
Real3DPortrait      & 20.88 & 0.780 & 0.154 & 0.801 & 0.150 & 0.268 & 5.971 & \colorbox{red!40}{0.663} & 0.296 & 0.411 \\
Portrait4D       & 20.35 & 0.741 & 0.191 & 0.765 & 0.144 & 0.205 & 4.854 & 0.596 & 0.286 & \colorbox{yellow!40}{0.258} \\
Portrait4D-v2   & 21.34 & 0.791 & 0.144 & 0.803 & 0.117 & 0.187 & 3.749 & \colorbox{yellow!40}{0.656} & 0.286 & 0.273 \\
GAGAvatar         & 21.83 & 0.818 & 0.122 & 0.816 & 0.111 & 0.135 & 3.349 & 0.633 &  0.253 & \colorbox{red!40}{0.247} \\
LAM              & \colorbox{yellow!40}{22.65} & \colorbox{yellow!40}{0.829} & 0.109 & 0.822 & \colorbox{yellow!40}{0.102} & \colorbox{yellow!40}{0.134} & \colorbox{yellow!40}{2.059}
                                   & 0.651 & \colorbox{yellow!40}{0.250} & 0.356 \\
\midrule
Ours~(Sub \#2)                  & \colorbox{red!40}{22.72} &                                          \colorbox{red!40}{0.831}
                                   & \colorbox{red!40}{0.091} &  \colorbox{red!40}{0.869} & \colorbox{red!40}{0.088} & \colorbox{red!40}{0.111} & \colorbox{red!40}{2.045}
                                   & \colorbox{orange!40}{0.660} & \colorbox{orange!40}{0.235} & \colorbox{orange!40}{0.257} \\
Ours~(Sub \#1)                  & \colorbox{orange!40}{22.68} &                                        \colorbox{orange!40}{0.830} 
                                   & \colorbox{orange!40}{0.094} & \colorbox{orange!40}{0.858} & \colorbox{orange!40}{0.089} & \colorbox{orange!40}{0.112} & \colorbox{orange!40}{2.055} & 0.644 & \colorbox{red!40}{0.233} &  0.260 \\
Ours~(Sub \#0)                  & 22.18 &  0.817 
                                   & \colorbox{yellow!40}{0.102} & \colorbox{yellow!40}{0.855} & 0.134 & 0.142 &  2.790 & 0.616 & 0.254 & 0.279 \\
\bottomrule
\end{tabular}}
\end{table*}

\begin{table*}[t]
\centering
\caption{Quantitative results on the HDTF dataset.  }
\label{tab:hdtf}

\resizebox{0.9\textwidth}{!}{%
\begin{tabular}{l|ccccccc|ccc}
\toprule
\multirow{2}{*}{\textbf{Method}} 
& \multicolumn{7}{c|}{\textbf{Self Reenactment}}
& \multicolumn{3}{c}{\textbf{Cross Reenactment}} \\
& \textbf{PSNR}$\uparrow$ & \textbf{SSIM}$\uparrow$
& \textbf{LPIPS}$\downarrow$ & \textbf{CSIM}$\uparrow$
& \textbf{AED}$\downarrow$ & \textbf{APD}$\downarrow$
& \textbf{AKD}$\downarrow$
& \textbf{CSIM}$\uparrow$ & \textbf{AED}$\downarrow$ & \textbf{APD}$\downarrow$ \\
\midrule
StyleHeat        & 21.41 & 0.785 & 0.155 & 0.657 & 0.158 & 0.162 & 4.585 & 0.632 & 0.271 & 0.239 \\
ROME                & 20.51 & 0.803 & 0.145 & 0.738 & 0.133 & 0.123 & 4.763 & 0.726 & 0.268 & 0.191 \\
OTAvatar          & 20.52 & 0.696 & 0.166 & 0.662 & 0.180 & 0.170 & 8.295 & 0.643 & 0.292 & 0.222 \\
HideNeRF          & 21.08 & 0.811 & 0.117 & 0.858 & 0.120 & 0.247 & 5.837 & 0.843 & 0.276 & 0.288 \\
GOHA                  & 21.31 & 0.807 & 0.113 & 0.725 & 0.162 & 0.117 & 6.332 & 0.735 & 0.277 & \colorbox{red!40}{0.136} \\
CVTHead          & 20.08 & 0.762 & 0.179 & 0.608 & 0.169 & 0.138 & 4.585 & 0.591 & 0.242 & 0.203 \\
GPAvatar         & 23.06 & 0.855 & 0.104 & 0.855 & 0.114 & 0.135 & 3.293 & 0.842 & 0.268 & 0.219 \\
Real3DPortrait     & 22.82 & 0.835 & 0.103 & 0.851 & 0.138 & 0.137 & 4.640 & \colorbox{red!40}{0.903} & 0.299 & 0.238 \\
Portrait4D      & 20.81 & 0.786 & 0.137 & 0.810 & 0.134 & 0.131 & 4.151 & 0.793 & 0.291 & 0.240 \\
Portrait4D-v2  & 22.87 & 0.860 & 0.105 & 0.860 & 0.111 & 0.111 & 3.292 & 0.857 & 0.262 & 0.183 \\
GAGAvatar        & 23.13 & 0.863 & 0.103 & 0.862 & 0.110 & 0.111 & 2.985 & 0.851 & 0.231 & 0.181 \\
LAM                             &  23.43 &\colorbox{yellow!40}{0.873} 
                                   & 0.097 & 0.865 & \colorbox{yellow!40}{0.101} & 0.093 & \colorbox{yellow!40}{1.965} & 0.849 & \colorbox{yellow!40}{0.230} & 0.229 \\
\midrule
Ours~(Sub \#2)                   & \colorbox{red!40}{24.14} &                                        \colorbox{red!40}{0.875}
                                   & \colorbox{red!40}{0.061} & \colorbox{red!40}{0.943} & \colorbox{red!40}{0.080} & \colorbox{red!40}{0.064} & \colorbox{red!40}{1.806}
                                   & \colorbox{orange!40}{0.886 }& \colorbox{red!40}{0.226} &\colorbox{orange!40}{0.155} \\
Ours~(Sub \#1)                  & \colorbox{orange!40}{24.06}     &\colorbox{orange!40}{0.874} 
                                       & \colorbox{orange!40}{0.063} & \colorbox{yellow!40}{0.937} & \colorbox{orange!40}{0.081} & \colorbox{orange!40}{0.066} & \colorbox{orange!40}{1.834} & \colorbox{yellow!40}{0.881} & \colorbox{orange!40}{0.227} & \colorbox{orange!40}{0.155}\\
Ours~(Sub \#0)                  & \colorbox{yellow!40}{23.85}  & 0.868
                                   & \colorbox{yellow!40}{0.067} & \colorbox{orange!40}{0.942} & 0.121 
                                   & \colorbox{yellow!40}{0.085} &  2.381 & 0.872 & 0.246 & 0.156\\
\bottomrule

\end{tabular}}%
\end{table*}

\begin{table*}[t]
  \centering
  \caption{The reenactment speed of neural-rendering-based methods measured in FPS, averaged over 100 frames. Driving parameters estimation time is excluded as they can be precomputed.}
  \vspace{-5pt}
  \label{table:fps}
  \renewcommand{\arraystretch}{1.0}
    \resizebox{\textwidth}{!}{  
  \begin{tabular}{c|cccccccc|ccc}
    \toprule
    & \multicolumn{8}{c|}{\textbf{A100 GPU}} & \multicolumn{3}{c}{\textbf{RTX 4090 GPU}} \\
    \textbf{Methods} & HideNeRF & CVTHead & Real3D & P4D-v2 & GAGavatar & Avat3r  & LAM & Ours~(Sub \#2) & Sub \#2 & Sub \#1 & Sub \#0 \\
    \midrule
    \textbf{FPS} & 9.73 & 18.09 & 4.55 & 9.62 & \colorbox{yellow!40}{67.12} & 53 & \colorbox{red!40}{280} &  \colorbox{orange!40}{85.94} & 126.44 & 148.04 & 152.57 \\
    \bottomrule
  \end{tabular}

}

\vspace{-10pt}
\end{table*}

\subsection{Quantitative Evaluation}
We report the quantitative results in ~\cref{tab:vfhq} and ~\cref{tab:hdtf}. Our method (Sub \#2) outperforms existing approaches across all reconstruction metrics (PSNR, SSIM, and LPIPS), as well as identity, expression, and pose consistency. Remarkably, our low-resolution LOD Sub \#1 surpasses LAM (80K Gaussians) and GAGAvatar (180K Gaussians) on both datasets  using only 29K Gaussian points, demonstrating the effectiveness of our hierarchical feature extraction and fusion strategy. 
We further report the inference efficiency in ~\cref{table:fps}. Our method achieves an inference speed of 85 FPS on an A100 GPU and 126 FPS on the consumer-grade RTX 4090 GPU, using the native PyTorch framework and the official implementation of 3D Gaussian Splatting. Compared to existing neural-rendering-based methods, our approach attains the highest inference speed. Moreover, our method outperforms LAM (280 FPS on A100 GPU without neural rendering) in terms of geometric details and dynamic textures, achieving an optimal balance between efficiency and visual quality.


\begin{figure}[t]
  \centering
  \resizebox{0.98\linewidth}{!}{
   \includegraphics{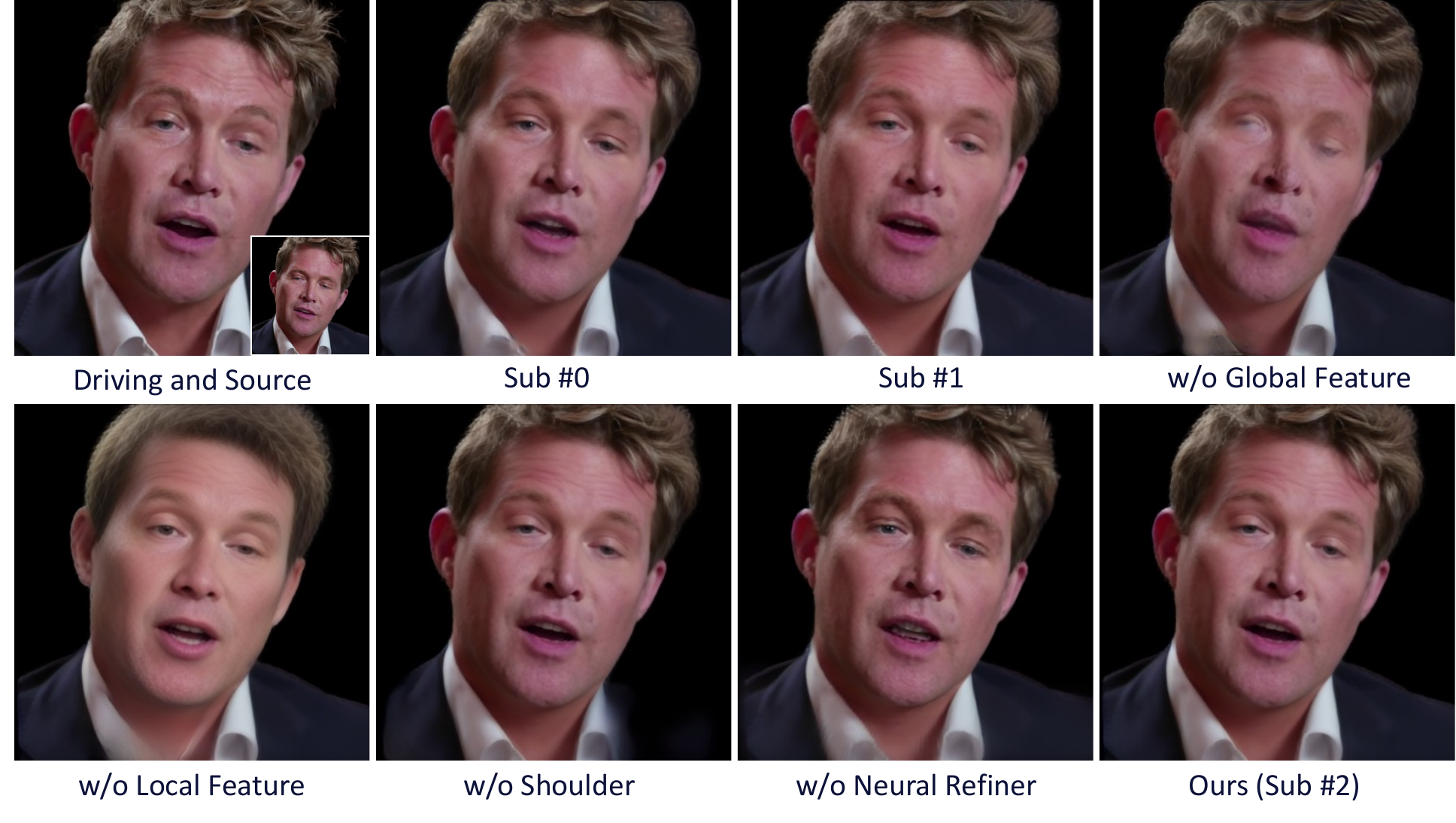}} 
       \vspace{-5pt}
  \caption{Our local-global feature fusion (OAFF) and multi-region fusion strategy significantly improve identity consistency and completeness in non-head regions. The neural refiner further boosts visual fidelity, especially for dynamic facial expressions.}
  \label{fig: ablations}
\end{figure}

\begin{table}[t]
  \centering
  \caption{Ablation results on the VFHQ dataset.}
  \vspace{-5pt}
  \small
            \begin{tabular}{l|cccc}
                    \toprule
            \textbf{Methods} & \textbf{PSNR}$\uparrow$ & \textbf{SSIM}$\uparrow$ & \textbf{LPIPS}$\downarrow$ & \textbf{CSIM}$\uparrow$ \\
            \midrule
            w/o Global Feature       & 20.85 & 0.796 & 0.121 & 0.429  \\ 
            w/o Local Feature       & 21.21 & 0.802 & 0.128 & 0.429  \\ 
            w/o Refiner    & 21.42 & 0.809 & 0.115  & 0.842 \\ 
            w/o Shoulder   & 22.42 & 0.828 & 0.099  & 0.867 \\ 
          
            \midrule
            Ours           & \textbf{22.72} & \textbf{0.831} & \textbf{0.091}  & \textbf{0.869}   \\ 
            \bottomrule
        \end{tabular}
        \label{tab:ablation}
 \vspace{-5pt}
\end{table}


          


\subsection{Ablation Studies}

\begin{figure}[t]
\begin{center}
\includegraphics[width=0.98\linewidth]{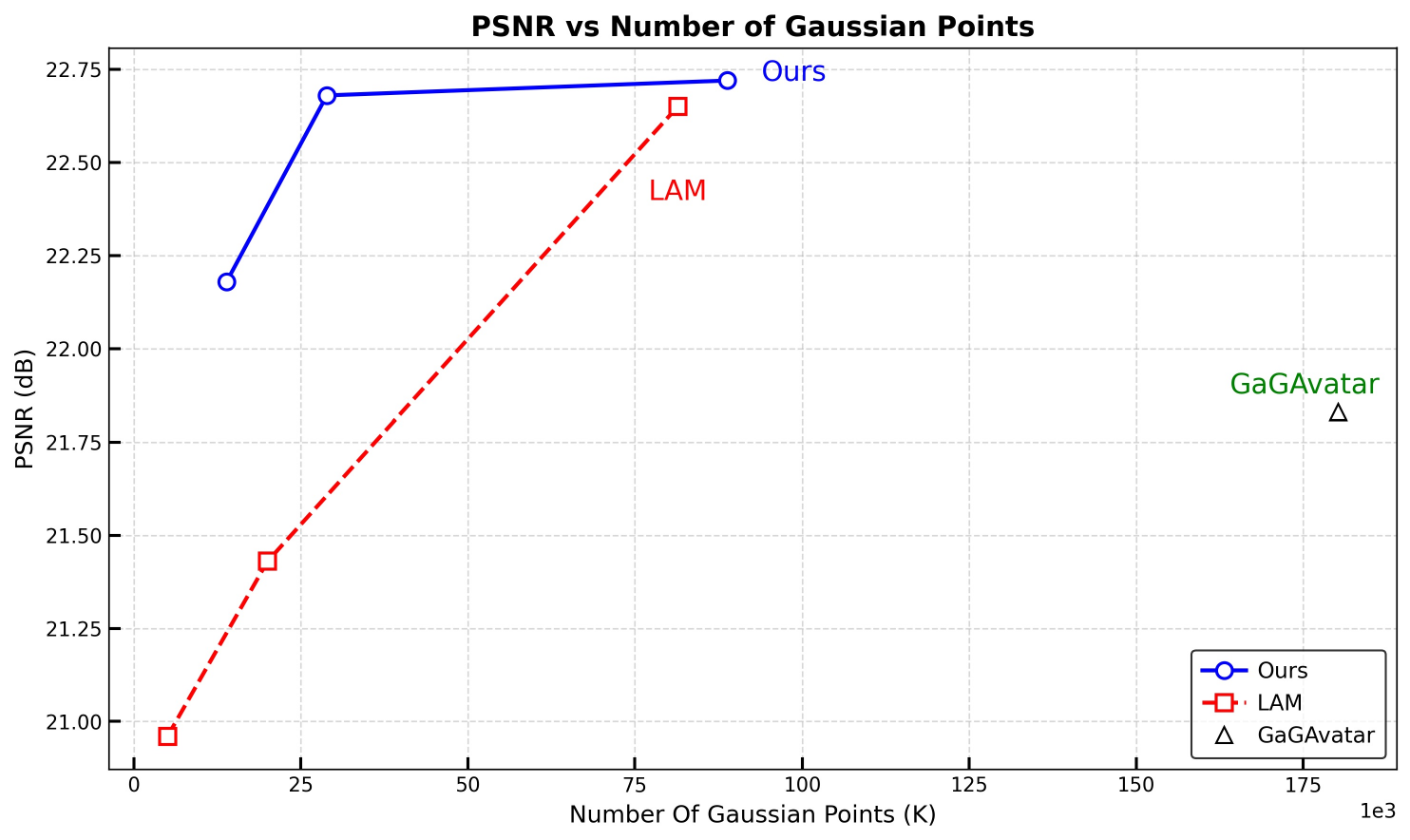}
\end{center}
\vspace{-10pt}
\caption{The correlation between Gaussian count and reconstruction performance on the VFHQ dataset.   
}
\vspace{-5pt}
\label{fig: psnr_vs_gaussians}
\end{figure}

\noindent{\bf Subdivision Times.}
To evaluate the effect of subdivision levels, we compare the results of our model with varying subdivision levels. The visual results are presented in ~\cref{fig: ablations}. Our observations indicate that higher subdivision levels capture more high-frequency details, such as wrinkles, leading to improved reconstruction quality. Quantitative comparisons are provided in ~\cref{tab:vfhq} and ~\cref{tab:hdtf}. We observe that performance saturates after two levels of subdivision (~\cref{fig: psnr_vs_gaussians}). We believe this is primarily due to the limited resolution of DINOv2’s feature maps: projection-based sampling beyond 80K vertices fails to recover additional geometric detail from the 88K (296 × 296) feature grid. Higher-resolution features could unlock further improvements. Additionally, increasing the number of subdivisions enhances the reconstruction quality at the cost of reduced inference speed, as shown in ~\cref{table:fps}. Notably, even our low-resolution LOD demonstrates competitive performance compared to existing methods, highlighting the effectiveness of our framework.

\noindent{\bf Local-Global and Multi-Region Feature Fusion.}
We conduct an ablation study by removing the global-local feature fusion in the OAFF module and using only global features. As shown in~\cref{fig: ablations} and~\cref{tab:ablation}, The absence of sampled local features significantly degrades identity consistency, while the lack of global features introduces artifacts in dynamic regions—particularly the eyes and mouth—that are inconsistent with the source image. Furthermore, when the shoulder region is excluded during rendering, the results display an incomplete and blurry appearance of the shoulder, as illustrated in~\cref{fig: ablations}.

\noindent{\bf Neural Rendering.}
We evaluate the effectiveness of the neural refiner module.~\cref{tab:ablation} and~\cref{fig: ablations} demonstrate that the neural refiner enhances fine details such as teeth and plays a key role in capturing expression-dependent features, including forehead wrinkles during eyebrow raising.

\begin{figure}[t]
  \centering
  \resizebox{0.98\linewidth}{!}{
   \includegraphics{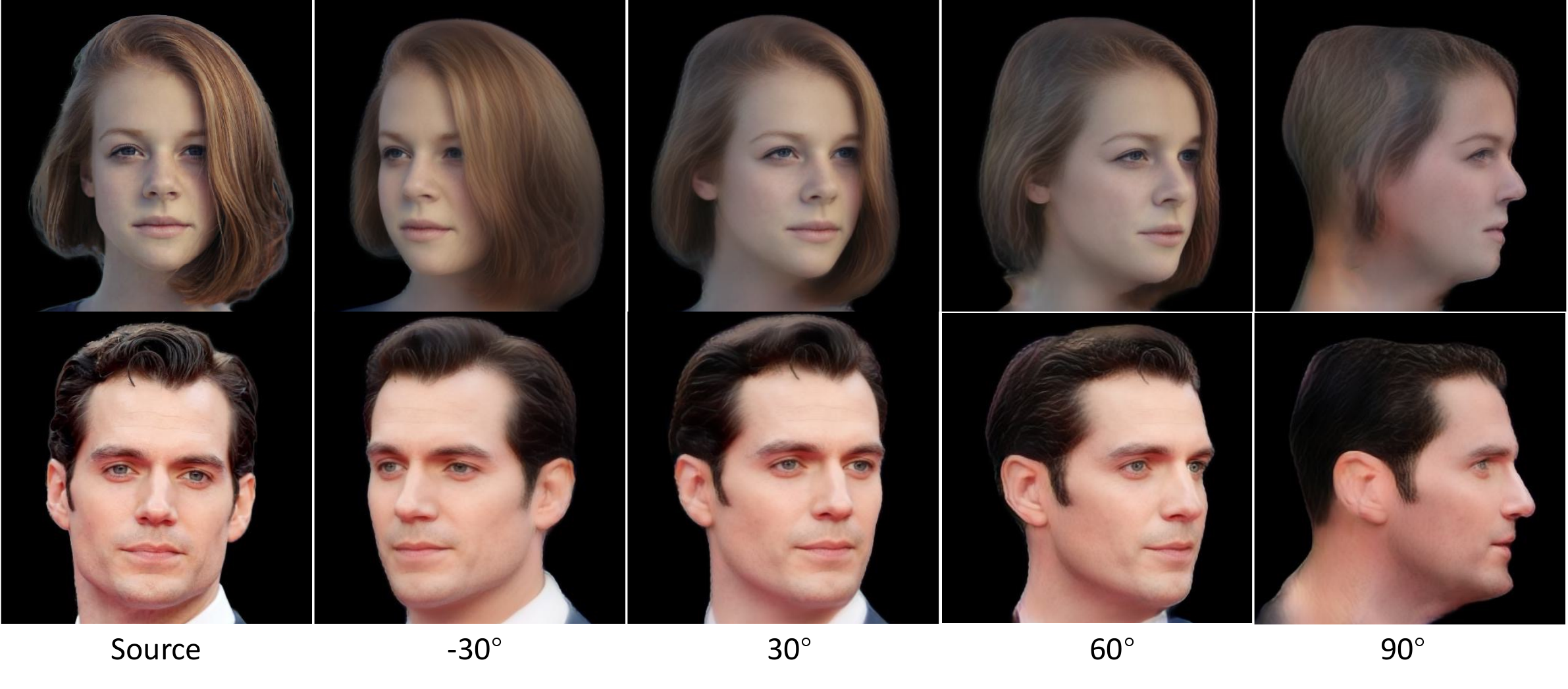}} 
       \vspace{-5pt}
  \caption{ Novel view results of OMG-Avatar. Since our model is trained on monocular interview videos, it achieves robust performance for head rotations within ± 60°, but exhibits noticeable artifacts beyond this range. }
  \label{fig:novelview}
  \vspace{-15pt}
\end{figure} 
\section{Conclusion}
In this paper, we present a novel multi-LOD framework for Gaussian head avatar reconstruction in a feed-forward manner. Our method enables dynamic level-of-detail (LOD) rendering at runtime, offering flexibility to accommodate varying device capabilities and inference speed requirements. Our model exhibits superior reconstruction and reenactment performance with significantly reduced computational cost. This is achieved by an efficient multi-level global and local feature extraction and a coarse-to-fine refinement strategy, as well as an occlusion-aware fusion mechanism. 
Moreover, our multi-region modeling scheme effectively enhances the visual fidelity of shoulder areas. 
Extensive experiments on two public datasets demonstrate that our approach outperforms state-of-the-art methods in terms of reconstruction quality, reenactment performance, and computational efficiency.

\noindent{\bf  Limitations and Future Work.}
Despite achieving strong results, our approach has two main limitations. First, the 3D Gaussian head model relies on the FLAME prior and accurate 3D morphable model (3DMM) tracking. However, FLAME does not capture fine facial dynamics—such as tongue motion, hair deformation, limiting the expressiveness of the generated avatars. Second, training solely on monocular videos reduces robustness to large viewpoint changes (\textgreater{} 60°, see the last column in~\cref{fig:novelview}). To address these issues, we plan to incorporate multi-view datasets for training, which will enhance spatial understanding and improve robustness across varying viewpoints.

\clearpage
{
    \small
    \bibliographystyle{ieeenat_fullname}
    \bibliography{main}
}

\clearpage
\setcounter{page}{1}
\maketitlesupplementary


\section{More Visualization Results}
\begin{figure*}[t]
\begin{center}
\includegraphics[width=0.9\linewidth]{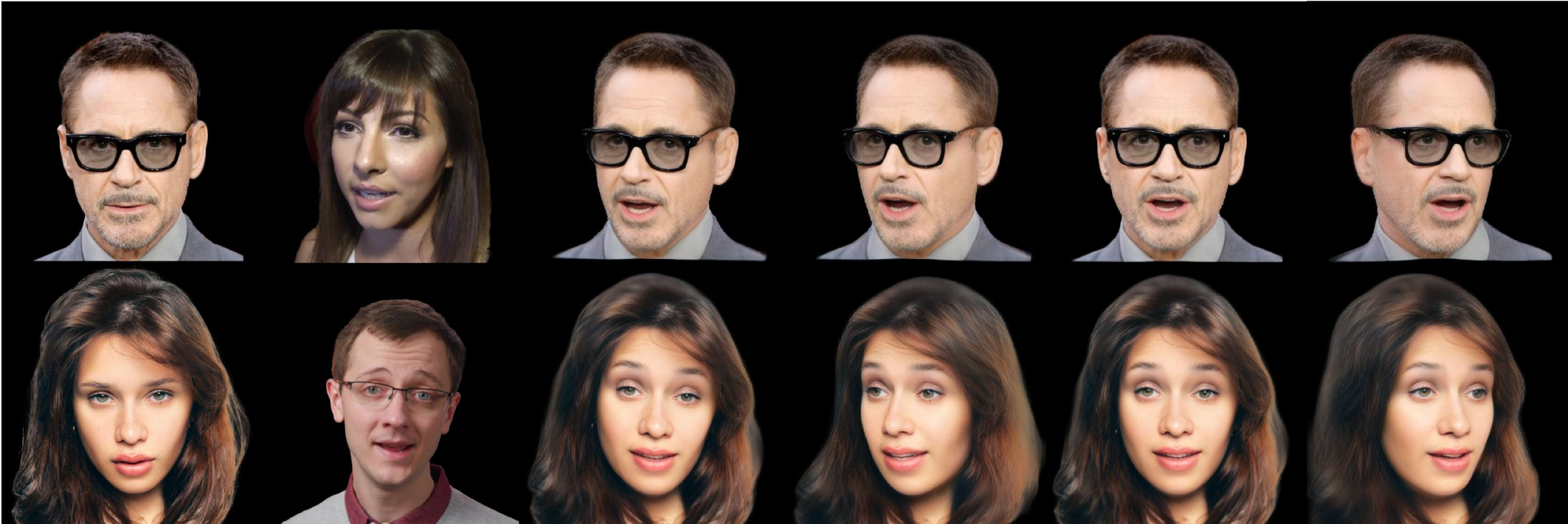}
\end{center}
\vspace{-5pt}
\caption{Reenactment and multi-view results of OMG-Avatar on in-the-wild images.}
\vspace{-15pt}
\label{fig:more_novel_view}
\end{figure*}

\begin{figure*}[t]
\begin{center}
\adjustbox{trim=0 0.03\height{} 0 0, clip}{
\includegraphics[width=0.99\linewidth ]{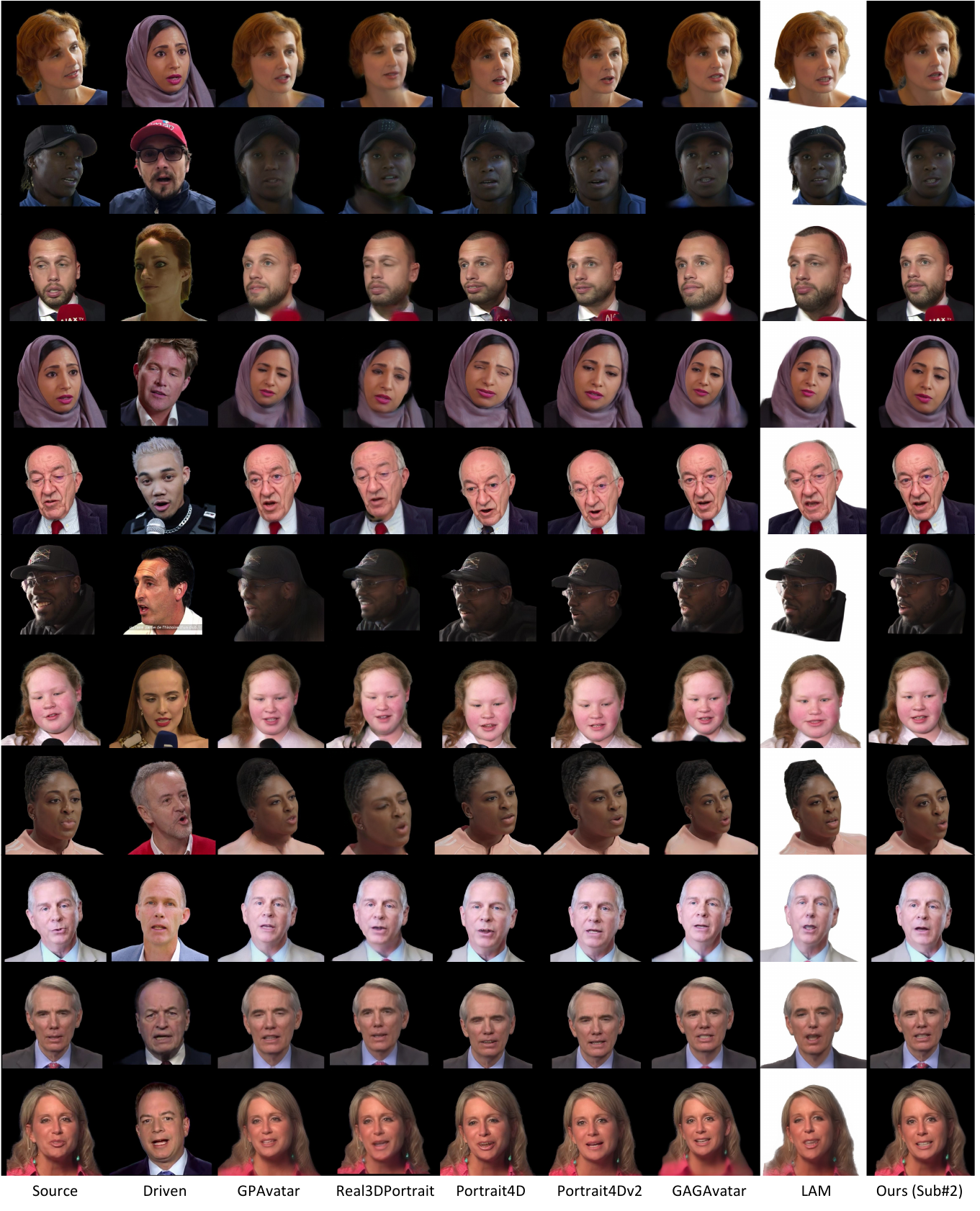}}
\medskip 
\parbox{0.99\linewidth}{%
  \centering
  \footnotesize
  \makebox[0.13\linewidth][c]{Source}%
  \makebox[0.08\linewidth][c]{Driving}%
  \makebox[0.13\linewidth][c]{GPAvatar}%
  \makebox[0.11\linewidth][c]{Real3DPortrait}%
  \makebox[0.13\linewidth][c]{Portrait4D}%
  \makebox[0.08\linewidth][c]{Portrait4Dv2}%
  \makebox[0.13\linewidth][c]{GAGAvatar}%
  \makebox[0.09\linewidth][c]{LAM}%
  \makebox[0.12\linewidth][c]{Ours(Sub\#2)}%
}

\end{center}
\caption{More cross-identity reenactment results on VFHQ and HDTF datasets.}
\vspace{-15pt}
\label{fig:cmp_more_cross}
\end{figure*}

\begin{figure*}[t]
\begin{center}
\includegraphics[width=0.95\linewidth]{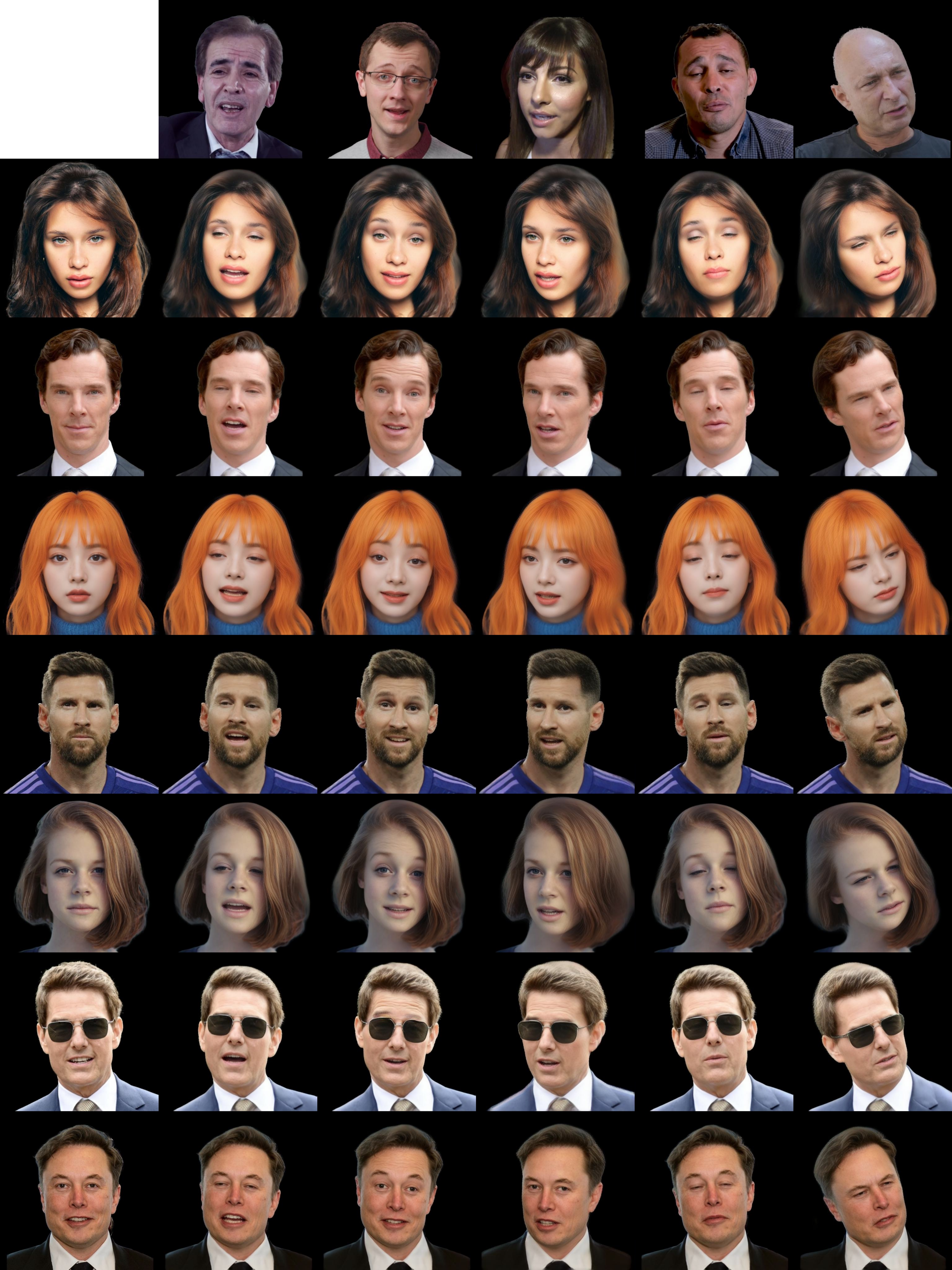}
\end{center}

\caption{Visualization of cross-reenacted results on in-the-wild images.}
\label{fig:wild1}
\end{figure*}

\begin{figure*}[t]
\begin{center}
\includegraphics[width=0.95\linewidth]{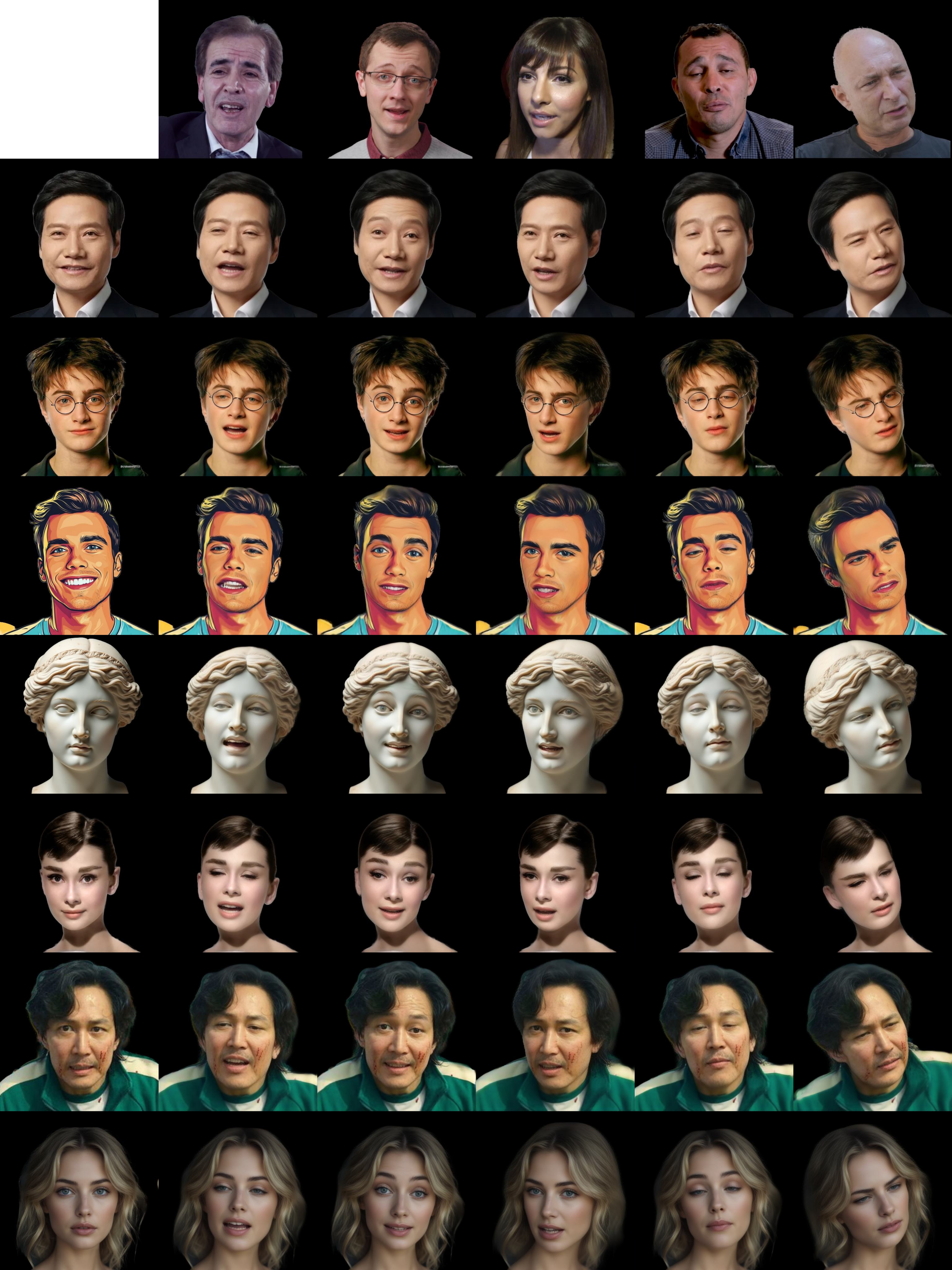}
\end{center}

\caption{Visualization of cross-reenacted results on in-the-wild images.}
\label{fig:wild2}
\end{figure*}

We present additional cross-reenactment results on the VFHQ and HDTF datasets in~\cref{fig:cmp_more_cross}, along with further qualitative evaluations on in-the-wild images in~\cref{fig:wild1} and~\cref{fig:wild2}. Novel view results are shown in~\cref{fig:more_novel_view}. Notably,~\cref{fig:wild2} demonstrates our model’s strong generalization capability—even to highly stylized or non-photorealistic inputs such as cartoon portraits (4th row) and statue-like sculptures (5th row). Our method faithfully preserves fine structural details in the shoulder region while accurately transferring the target pose and expression. This is exemplified by the clear rendering of the bloodstain on the collar in the second-to-last row of~\cref{fig:wild2}. Furthermore, our approach exhibits remarkable robustness to facial occlusions—such as sunglasses in the second-to-last row of~\cref{fig:wild1}—without degrading reenactment quality or introducing visible artifacts. 

To the best of our knowledge, GAGAvatar~\cite{chu2024generalizable}, LAM~\cite{he2025lam} and ours are the only state-of-the-art methods that support one-shot, feed-forward 3D avatar reconstruction with real-time facial reenactment. We compare our approach against both and present the results in~\cref{fig:cmp_sota_realtime}. Additional video reenactment results on in-the-wild imagery are provided in the supplementary video (OMG-Avatar.mp4).

\section{Ethical Discussion}
Our method enables high-fidelity, animatable 3D head avatar generation with potential applications in video production, digital communication, and other domains. However, like other advanced generative models, it could be misused to create deceptive or non-consensual synthetic content (commonly known as "deepfakes") that may mislead, manipulate, or infringe on personal privacy. We firmly oppose such misuse and emphasize that our work is intended solely for legitimate, consent-based applications. To mitigate potential risks, we propose the following safeguards:

\begin{figure}[h]
\begin{center}
\includegraphics[width=0.99\linewidth]{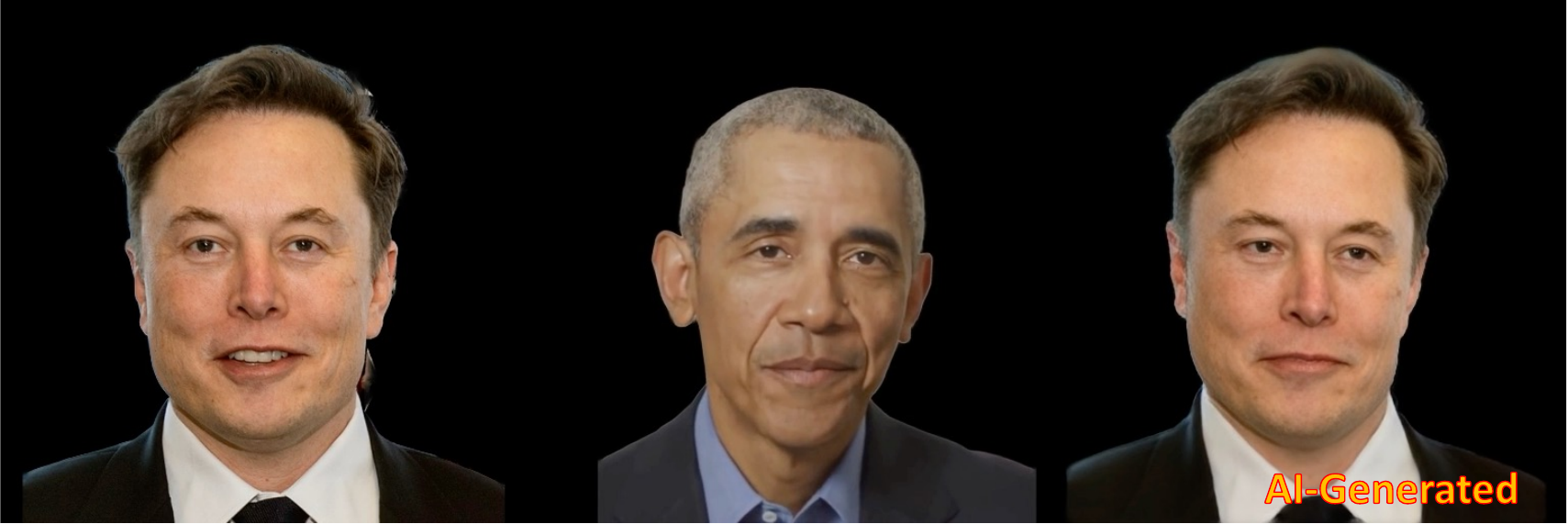}
\end{center}
\caption{Visible watermarks will be embedded in all generated images and videos, clearly indicating that the content is AI-generated.}

\label{fig:watermark}
\end{figure}

\begin{itemize}
\item \textbf{Visible and invisible watermarking.} We will integrate visible watermarking mechanisms into our released code, as shown in Figure~\ref{fig:watermark}. Visible watermarks will be embedded in all generated images and videos, clearly indicating that the content is AI-generated, enabling viewers to easily distinguish synthetic media from authentic recordings. In addition, we plan to adopt robust invisible watermarking techniques~\citep{tancik2020stegastamp} that embed and reliably decode arbitrary data (\emph{e.g.}, hyperlinks) in a perceptually invisible manner, while remaining resilient to real-world distortions such as compression, printing, and re-photography. These watermarks are designed to be difficult to remove without degrading visual quality.


\item \textbf{Strict licensing.} Our code and models will be released under a restrictive license that prohibits the creation of avatars based on real individuals without explicit consent, particularly for commercial purposes. The license further restricts usage to ethical and non-deceptive applications, and any violation can be traced through the embedded watermarking system.
\end{itemize}

While our method advances the state of animatable head generation, we acknowledge its dual-use potential. Through technical measures like watermarking and policy-level controls via licensing, we aim to minimize the risk of abuse. As technology developers, we have the responsibility to build safeguards into our systems. However, preventing misuse requires broader efforts, from platform policies, legal frameworks, and user awareness. We call on researchers, developers, and content creators to exercise ethical judgment and social responsibility when deploying generative avatar systems. With appropriate oversight and responsible use, our work can contribute positively to immersive communication and other beneficial applications.


\section{Reproducibility Details}
\label{app:Reproducibility}


\begin{table*}[h]
    \centering
    \begin{tabular}{c|ccc}
        \toprule
        \textbf{Subdivision Times} & \textbf{Head Only}  & \textbf{Head+Shoulder on VFHQ}  & \textbf{Head+Shoulder on HDTF} \\
       
        \midrule
        \textbf{\#0} & 5, 023 & 13, 883 & 14, 466  \\ 
        \textbf{\#1} & 20, 018 & 28, 878 & 29, 461 \\ 
        \textbf{\#2} & 79, 936 & 88, 796 & 89, 379  \\ 
          \midrule
         \textbf{Shoulder Points}  &   &  +8, 860 & +9, 443  \\ 
        \bottomrule
    \end{tabular} 
    \caption{ The number of Gaussian points for different subdivision level. Integrating the shoulder region leads to an average increase of 8, 860 Gaussian points on the VFHQ dataset and 9, 443 on the HDTF dataset.}
    \label{tab: subdivision_count}
\end{table*}

\begin{table*}[h]
    \centering
      \resizebox{0.8\textwidth}{!}{
    \begin{tabular}{@{}l|ccccc@{}}
        \toprule
        & \textbf{\# Transformer} & \textbf{\# Attention Heads} & \textbf{Attention} & \textbf{Feature} & \textbf{Training} \\
        & \textbf{Layers} & \textbf{Per Layer} & \textbf{Map Size} & \textbf{Dimension} & \textbf{GPU Hours} \\ 
        \midrule
        \textbf{LAM} & 10 & 16 & $80\text{k} \times N_{\text{DINO}}$ & 1024 & $\sim 2600\,\text{h}$ (2 weeks $\times$ 8 GPUs) \\ 
        \textbf{Ours} & 2 & 8 & $5\text{k} \times N_{\text{DINO}}$ & 256 & $\sim 200\,\text{h}$ \\ 
        \bottomrule
    \end{tabular}}
    \caption{Comparison of training configurations for LAM and ours.}
    \label{tab: computation_comparison}
\end{table*}

\subsection{Dataset Processing}
We construct our training and testing datasets by uniformly sampling frames from the videos in the VFHQ dataset. Specifically, we use 15, 204 video clips for training and 50 clips for testing. For the training set, we sample a number of frames $ N $ per clip based on the video length, following the strategy proposed in~\citet{chu2024generalizable}:

\begin{itemize}
    \item $ N = 25 $ if the video length is less than 200 frames,
    \item $ N = 50 $ if the video length is between 200 and 300 frames,
    \item $ N = 75 $ if the video length exceeds 300 frames.
\end{itemize}
This results in a total of 766, 263 training frames. For testing, we sample $ N = 50 $ frames per video, yielding 2, 500 test frames in total.

To evaluate the generalization ability of our model, we directly test it on the HDTF dataset using weights trained on the VFHQ dataset. We follow the dataset split setting from \citet{ma2023otavatar}, and uniformly sample 100 frames per video, resulting in a total of 1, 900 frames for evaluation.

\subsection{More Implementation Details}
We utilize a frozen DINOv2 model to extract both local and identity features from an input image.The local feature has a size of $256 \times 296 \times 296$, while the identity feature is of size $1369 \times 768$. The original FLAME mesh contains 5, 023 vertices, and its positional encoding has a size of 5, 023 $\times$ 256. After passing through a two-layer Transformer, we obtain a global feature map with dimensions 5, 023 $\times$ 256. Subdividing the global feature results in hierarchical representations at level 1 and level 2, with sizes of 20, 018 $\times$ 256 and 79, 936 $\times$ 256, respectively. The first dimension of these features corresponds to the number of vertices at each subdivision level. 

During training, the number of subdivisions is progressively increased according to the training progress. Specifically, no subdivision is applied in the early phase ($\leq 10\%$ of total iterations), one level is used in the intermediate phase (10\%--30\%), and a random strategy with a bias towards 2 (70\% for 2, 20\% for 1, and 10\% for 0) in the later stage. The model is trained for 6 epochs on a single NVIDIA A100 GPU using the Adam optimizer and a linear learning rate decay schedule, with an initial learning rate of $1 \times 10^{-4}$ and a batch size of 8.


\section{Mesh Subdivision}
\label{sec: Subdivision}
We apply the Loop subdivision algorithm~\citep{loop1987smooth} to perform mesh subdivision on the FLAME model, utilizing its implementation in PyTorch3D. The Loop subdivision algorithm refines a triangle mesh by introducing a new vertex at the midpoint of each edge and dividing each triangular face into four smaller triangles. Additionally, vertex attribute vectors (\emph{e.g.}, $F^{GS}_{global}$ in~\cref{eq: subdivider} ) are subdivided by averaging the attribute values of the two vertices that form each edge. 

In Table~\ref{tab: subdivision_count}, we present the number of Gaussian points for different subdivision levels, and the shoulder region approximately adds 9K additional points.

\section{Computational Cost Analysis}
Similar to our approach, the LAM method also utilizes mesh subdivision to increase the number of points, thereby improving the Gaussian's capability to capture fine-grained details. Both methods set the number of subdivision iterations to 2. However, the computational complexity differs significantly. Let $k$ denote the subdivision level. After $k$ iterations, the number of vertices in the mesh is approximately $4^kV_0$, where $V_0$ represents the number of vertices in the original FLAME mesh. Consequently, the computational complexity of the most expensive module, the cross attention, is:
\begin{equation}
O(l \cdot h \cdot 4^k \cdot V_0 \cdot N_\text{DINO} \cdot d_\text{head}) 
\end{equation}
where $ l $ denotes the number of transformer layers, $ h $ is the number of attention heads, $ N_{\text{DINO}} $ is the number of DINO features, and $ d_{\text{head}} $ is the feature dimension. In our method, cross-attention is computed at the 0-th level (before any subdivision), while LAM performs it on the finest-level subdivided mesh, where the computational complexity grows exponentially with the number of subdivisions $ k $. As a result, the reconstruction cost of our method is only $ 1/640 $ that of LAM. As shown in~\cref{tab: computation_comparison}, our training GPU-hours are more than 90\% lower than LAM’s.

\begin{figure*}[t]
\begin{center}
\includegraphics[width=0.9\linewidth]{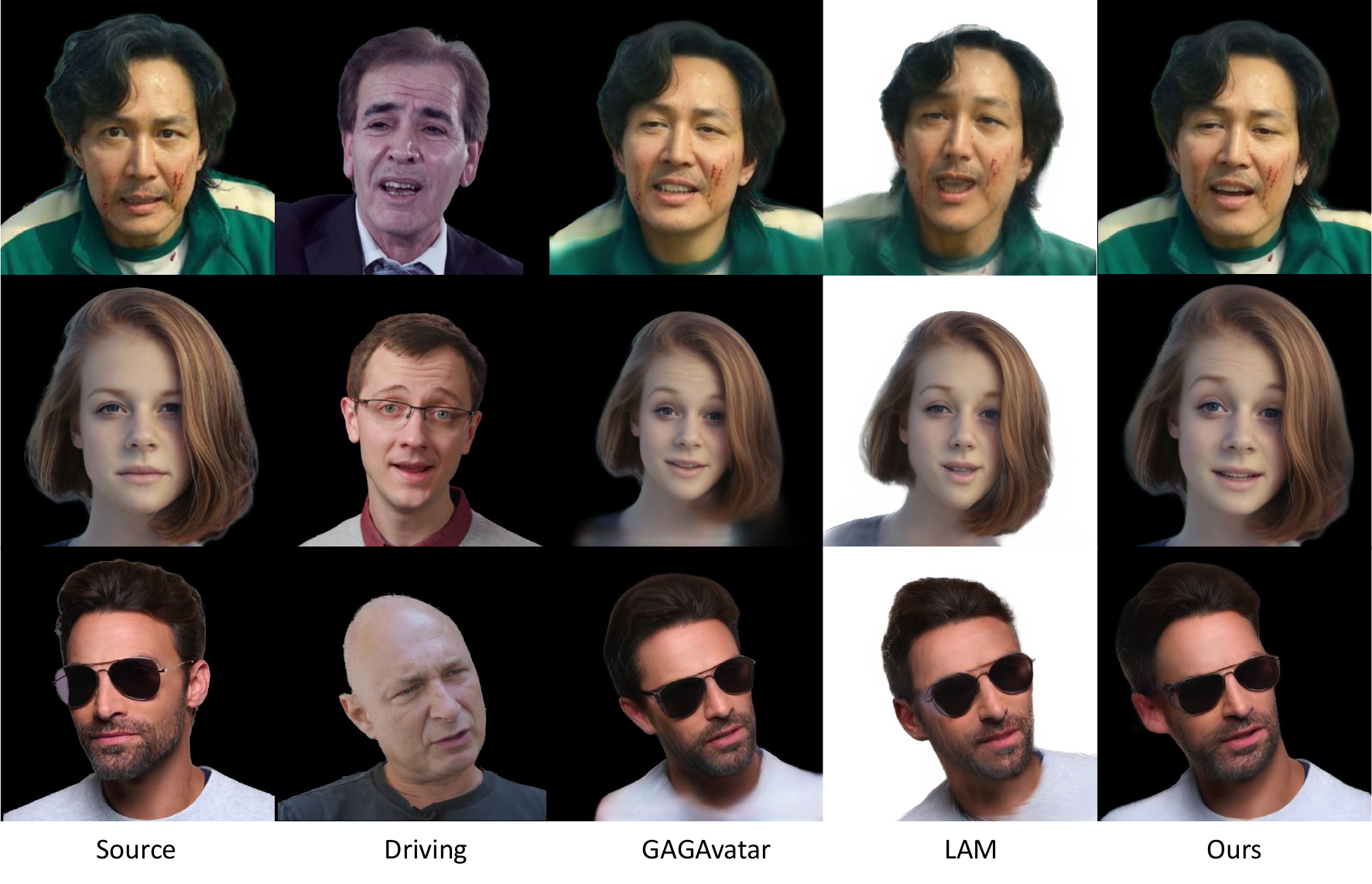}
\end{center}
\vspace{-15pt}
\caption{Comparision with state-of-the-art methods that support one-shot, feed-forward 3D avatar reconstruction with real-time facial reenactment.}

\label{fig:cmp_sota_realtime}
\end{figure*}

\end{document}